\documentclass{article}

\usepackage{microtype}
\usepackage{graphicx}
\usepackage{subcaption}
\usepackage{booktabs}
\usepackage{array}
\usepackage{amsmath}
\usepackage{amssymb}
\usepackage{mathtools}
\usepackage{fancyvrb}
\usepackage{hyperref}

\usepackage[preprint]{icml2026}
\usepackage[capitalize,noabbrev]{cleveref}

\newcommand{\Lean}{Lean}
\newcommand{\mathlib}{mathlib}
\newcommand{\code}[1]{\texttt{\detokenize{#1}}}
\newcommand{\codesmall}[1]{{\scriptsize\texttt{\detokenize{#1}}}}
\newcolumntype{L}[1]{>{\raggedright\arraybackslash}p{#1}}

\icmltitlerunning{Sorries Are Not the Hard Part}

\icmlsetsymbol{equal}{*}

\begin{document}

\twocolumn[
  \icmltitle{Sorries Are Not the Hard Part:\\
  An Expert-Review Case Study of a Semi-Autonomous Formalization}

  \begin{icmlauthorlist}
    \icmlauthor{Vasily Ilin}{equal,uw}
    \icmlauthor{Brian Nugent}{equal,utah}
  \end{icmlauthorlist}

  \icmlaffiliation{uw}{University of Washington, Seattle, WA, USA}
  \icmlaffiliation{utah}{University of Utah, Salt Lake City, UT, USA}
  \icmlcorrespondingauthor{Vasily Ilin}{vasilin97@gmail.com}
  \icmlkeywords{autoformalization, theorem proving, Lean, mathematical software engineering}

  \vskip 0.3in
]

\printAffiliationsAndNotice{\icmlEqualContribution}

\begin{abstract}
Large language models can often close proof gaps in interactive theorem
provers, but a verified theorem is not the same thing as a reusable library
contribution.  We study this distinction through a detailed case study:
a semi-autonomous formalization of Grothendieck's
vanishing theorem. The initial version compiles with no sorries, but an expert review
found serious problems in definitions, theorem generality, file organization,
and the API. We then ran a review-driven refactor and compression
process and obtained a second expert review. The before-and-after comparison
shows a sharp split: agents adapted well to local, mechanically checkable
feedback, but remained weak at choosing definitions and designing APIs.  We
argue that autoformalization should be evaluated not only by closed sorries,
but by whether the resulting formalization survives expert review.
\end{abstract}

\section{Intro}

The default public story for semi-autonomous formalization is proof
completion: given an informal or formal statement, can an agent produce code
accepted by a proof assistant?  This is the right first question, but it is
not the last.  Modern mathematics is highly interconnected, and a formal
library that supports it must compose: definitions must combine, theorem
statements must have the right generality, namespaces must support search,
and APIs must let future users reason about properties rather than unfold
implementations.

This paper evaluates the quality of LLM-produced Lean code and asks what
the current barriers are between it and a reusable library contribution.
Our case study is a \Lean{} formalization of Grothendieck's vanishing
theorem \citep[Ch.~III, Thm.~2.7]{hartshorne1977}: if \(X\) is a Noetherian
topological space and \(n\) is above the topological Krull dimension of
\(X\), then the \(n\)-th sheaf cohomology of any sheaf of abelian groups on
\(X\) vanishes. The authors wrote the Lean statement by hand, supplied
Claude Code with a PDF excerpt of Hartshorne's proof, and instructed the
agent to follow that proof.  The statement uses only existing \mathlib{}
definitions:
\begin{Verbatim}[fontsize=\small]
theorem GrothendieckVanishing
    (X : TopCat) [NoetherianSpace X]
    (n : Nat) (h : n > topologicalKrullDim X)
    (F : Sheaf AddCommGrpCat X) :
    Subsingleton (Sheaf.H F n)
\end{Verbatim}
That the statement uses only existing definitions matters: it prevents the
agent from making custom definitions to render the theorem trivially true.

\paragraph{Proof sketch.}
The formal proof follows Hartshorne's route by induction on
\code{topologicalKrullDim X}.  First it reduces to the irreducible case
using a closed-immersion short exact sequence and the fact that Noetherian
spaces have finitely many irreducible components.  For an irreducible space
of dimension zero, every open is simple enough that any sheaf is flasque,
and higher cohomology of any flasque sheaf vanishes.  In the
positive-dimensional irreducible case, the proof chooses a proper
irreducible closed \(Z \subset X\), uses strict dimension drop on \(Z\),
and combines the induction hypothesis with closed-immersion cohomology,
extension by zero, and a finite-generation reduction through filtered
colimits.  These reductions need supporting definitions and API not
currently in \mathlib{} --- for the closed-immersion short exact sequence,
flasque vanishing, extension by zero, finite-generation via filtered
colimits, topological Krull dimension, and sheaf cohomology --- which
serve as our tests for the agent's ability to define new objects and build
around them.

\paragraph{Before and After.}
We treat the project as a before-and-after experiment.  State A is the
first verified version, inspected by one of the authors, a
\Lean{}/\mathlib{} expert.  State B is the version after an automated
response to that review and a later mathlib-style cleanup, inspected by
the same expert.  The review exposes what proof-completion metrics miss:
LLMs are much better at closing local goals than at deciding what objects
should exist.

The contributions are:
\begin{itemize}
  \item a case study of semi-autonomous formalization of a graduate-level
  algebraic geometry theorem in \Lean{};
  \item a qualitative before-and-after analysis of expert review feedback on
  definitions, theorem statements, proofs, and file/API structure.
\end{itemize}

\paragraph{Why this theorem is a useful stress test.}
Grothendieck vanishing is not a contest-style problem reducible to a single
tactic script.  The theorem is short only because it sits on a tower of
categorical and topological infrastructure: the agent had to work
simultaneously with presheaves, sheaves, stalks, exact sequences, filtered
colimits, Krull dimension, and the derived-category definition of
cohomology.  The central claim is deliberately modest: we do not claim
one formalization predicts all future work, only that kernel acceptance is
an incomplete evaluation target, and that this case shows why.

\section{Related Work}

\paragraph{Formal libraries as shared interfaces.}
The case study builds on \Lean{} \citep{demoura2015lean,ullrich2021lean4}
and \mathlib{}, a single shared library supporting modern mathematics at
scale \citep{mathlib2020}.  Evaluating a contribution to such a library is
closer to code review than to grading a fixed benchmark theorem.

\paragraph{LLMs for proof generation.}
Autoformalization and neural theorem proving have largely been framed as
translation \citep{wu2022autoformalization}, retrieval, and proof-search
problems: LeanDojo \citep{yang2023leandojo} and Baldur \citep{first2023baldur}
target retrieval-augmented proving and whole-proof repair, APRIL
\citep{wang2026april} learns proof repair and iterative refinement from
compiler feedback, and recent
systems---Seed-Prover \citep{seedprover2025}, Aristotle \citep{aristotle2025},
Numina-Lean-Agent \citep{numina2026}, and QED-Nano \citep{qednano2026}---scale
proof search through formal feedback, agentic tool use, and distillation.  A
few efforts push past binary correctness: \citet{proofrank2026} score whether
a proof is clear, concise, and transferable, \citet{mathlibpr2026} test
whether models can judge \mathlib{} pull-request merge-readiness, and
\citet{klingner2026eval} evaluate formalization in realistic imported contexts
(miniF2F, miniCTX).  In all of these the unit of evaluation is the attempted
theorem; ours is the library delta that survives after it closes --- the
duplicated definition or over-specific lemma a theorem benchmark has no reason
to penalize.

\paragraph{Research-scale semi-autonomous formalization.}
Several recent projects move beyond contest problems: rapid large-scale
topology \citep{urban2026topology}, parallel Diophantine formalization in
Isabelle \citep{davidbayer2025}, a Lean/PhysLean formalization that uncovered
a literature error \citep{toobysmith2026}, autonomous number-theory and
algebra formalizations \citep{axiom2026fel,axiom2026partial}, Archon's
previously open commutative-algebra conjecture in roughly 19{,}000 lines
\citep{archon2026}, LeanMarathon's multi-agent construct--audit--prove--repair
harness over four Erd\H{o}s problems \citep{leanmarathon2026}, and AutoformBot's
thousands of agents formalizing 26 textbooks into over 45{,}000 declarations
\citep{rammal2026scale}.  Closest to our setting, the Vlasov-Maxwell-Landau
case study \citep{ilin2026vml} also emphasizes full-process logs, agent
failure modes, and human review of key statements.  These reports show AI can
produce substantial verified artifacts quickly and that process logs expose
failures a final proof hides.

\paragraph{What the prior work leaves open.}
What none of this isolates is the gap between a verified artifact and a
reusable one.  Proof-generation evaluations have an unambiguous terminal
condition --- the artifact checks or it does not --- but for library
construction the statement and its intermediate abstractions are themselves
outputs, and a proof can pass the kernel while forcing future users through
awkward coercions or duplicated definitions.  Large-scale reports include rich
process data, but their notion of quality is still dominated by completion.
We instead make the expert review the object of study, letting us ask which
complaints an automated loop fixes and which persist because they require
judgment about future reuse rather than proof search.

\section{Timeline}

\begin{figure*}[t]
  \centering
  \includegraphics[width=.92\textwidth]{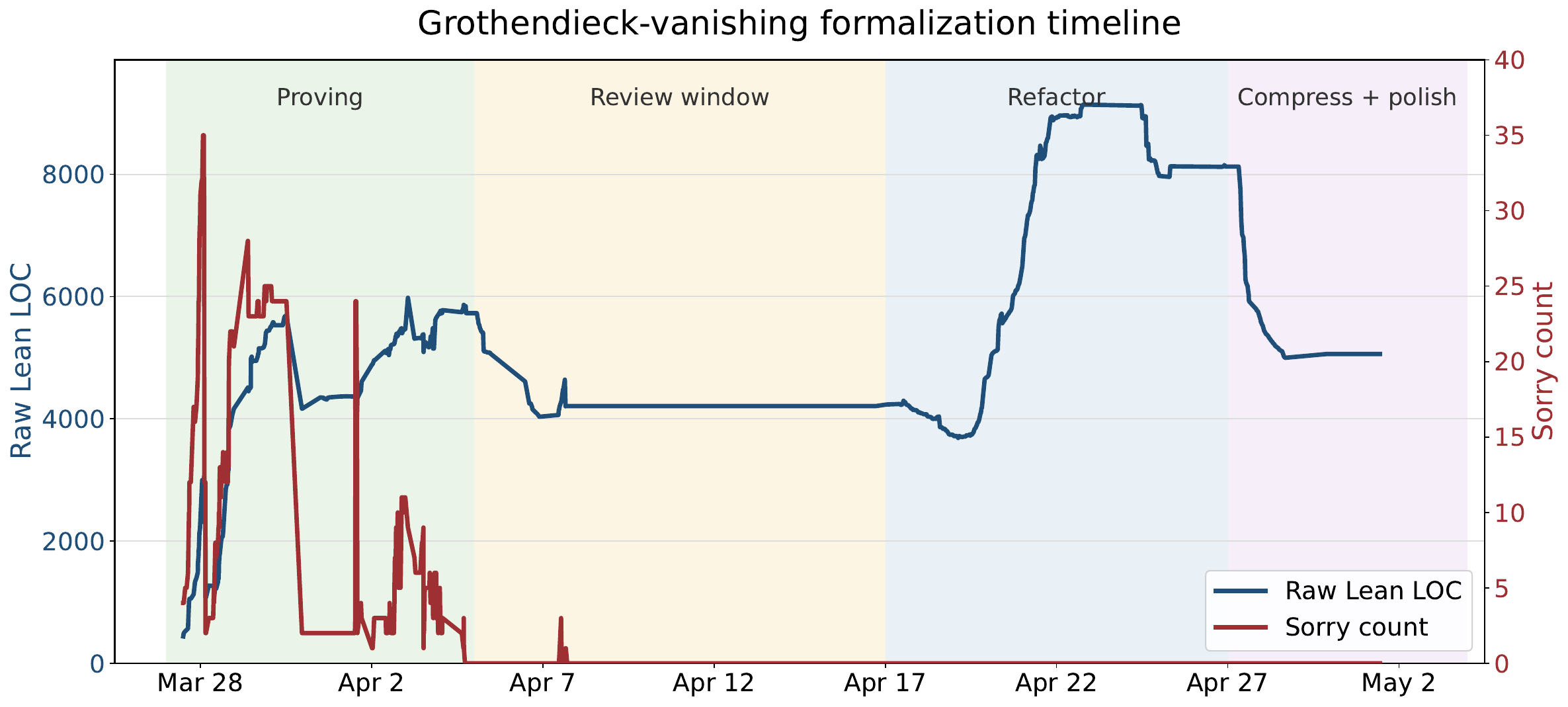}
  \caption{Project timeline reconstructed from the commit log.  The blue
  curve is raw \Lean{} lines of code and the red curve is the sorry count.
  Background colors mark the process phases: green = proving,
  yellow = expert-review window, blue = review-driven refactor,
  purple = compression and final mathlib-style polish.  The proving phase
  produces the first sorry-free version by April~4; LOC then keeps moving
  well after the sorry count reaches zero, as the project is rewritten in
  response to the expert review.}
  \label{fig:timeline}
\end{figure*}

\paragraph{Reconstructing the project.}
\Cref{fig:timeline} and \cref{tab:phase-timeline} reconstruct the project
from its commit history, structured loop logs, telemetry summaries, and
two expert reviews of the same development.\footnote{The two expert
reviews and the structured process logs analyzed in this paper are
released as a public dataset:
\url{https://huggingface.co/datasets/uw-math-ai/grothendieck-vanishing-logs}.
The \Lean{} source is available at
\url{https://github.com/Vilin97/Clawristotle/tree/main/grothendieck-vanishing}.}  The mathematical statement and proof outline were supplied by humans.

\begin{table*}[t]
\caption{Project phases reconstructed from commits, loop logs, and reviews.}
\label{tab:phase-timeline}
\centering
\small
\begin{tabular}{L{0.16\textwidth}L{0.16\textwidth}L{0.28\textwidth}L{0.28\textwidth}}
\toprule
Phase & Duration & What happened & Main difficulty \\
\midrule
Formalization & Mar. 27--Apr. 4; nine active days &
Claude Code followed the supplied Hartshorne proof plan, with Aristotle used
for bounded lemmas.  This produced state A, the first sorry-free version. &
The agent could close local goals, but the emerging API was ad hoc.  A
heartbeat detour also stopped the project from compiling cleanly until large
proofs were decomposed. \\
Expert review & Apr. 8--15; one week &
The same expert read the first verified tree as library code and wrote a
structured audit with large cross-file changes and per-file comments. &
The review judged the code after kernel success, so it found design failures
that no proof-completion metric could see. \\
Review response & Apr. 17--May 1; about two weeks &
The main refactor loop ran Apr. 19--27. &
Local checklist items were tractable; open-ended requests such as building a
usable \code{Sheaf.H} API were much harder. \\
Compression and polish & Apr. 27--May 1 &
The compression loop ran Apr. 27--28, followed by a
mathlib-style cleanup pass for names, docstrings, lint, and stale comments. &
The LOC gate found real redundancy, but short code was only a proxy for
library quality. \\
\bottomrule
\end{tabular}
\end{table*}

\paragraph{The proving phase.}
The proving phase was fast because the route was fixed in advance.  The
sorry curve in \cref{fig:timeline} is nonetheless jagged, because the
agent scaffolded intermediate claims, discharged some, and repeatedly
changed the proof decomposition.  The clearest side quest is the
heartbeat episode (Mar.~28 -- Apr.~1).  Mathlib's \code{synthInstance}
budget for \code{HasDerivedCategory} kept colliding with proofs
mentioning \code{Ext} or \code{Sheaf.H}, and the agent's first instinct
was to raise \code{set_option maxHeartbeats}, oscillating budgets between
200K and 12.8M across days without producing a stable project.  The
episode only ended after one author told the agent to ``keep optimizing
until the whole project compiles with default heartbeats''.  Stripping
the overrides regressed the sorry count from 3 to 24; recovery came from
caching instances via \code{inferInstanceAs} and splitting large proofs
into named sub-lemmas.  The durable rule that came out of this is: never raise \code{maxHeartbeats} above
200000.  It is the first appearance of a pattern that recurs throughout
the project: a proof too expensive to check is usually a proof that
should be decomposed.

\begin{figure*}[t]
  \centering
  \includegraphics[width=.95\textwidth]{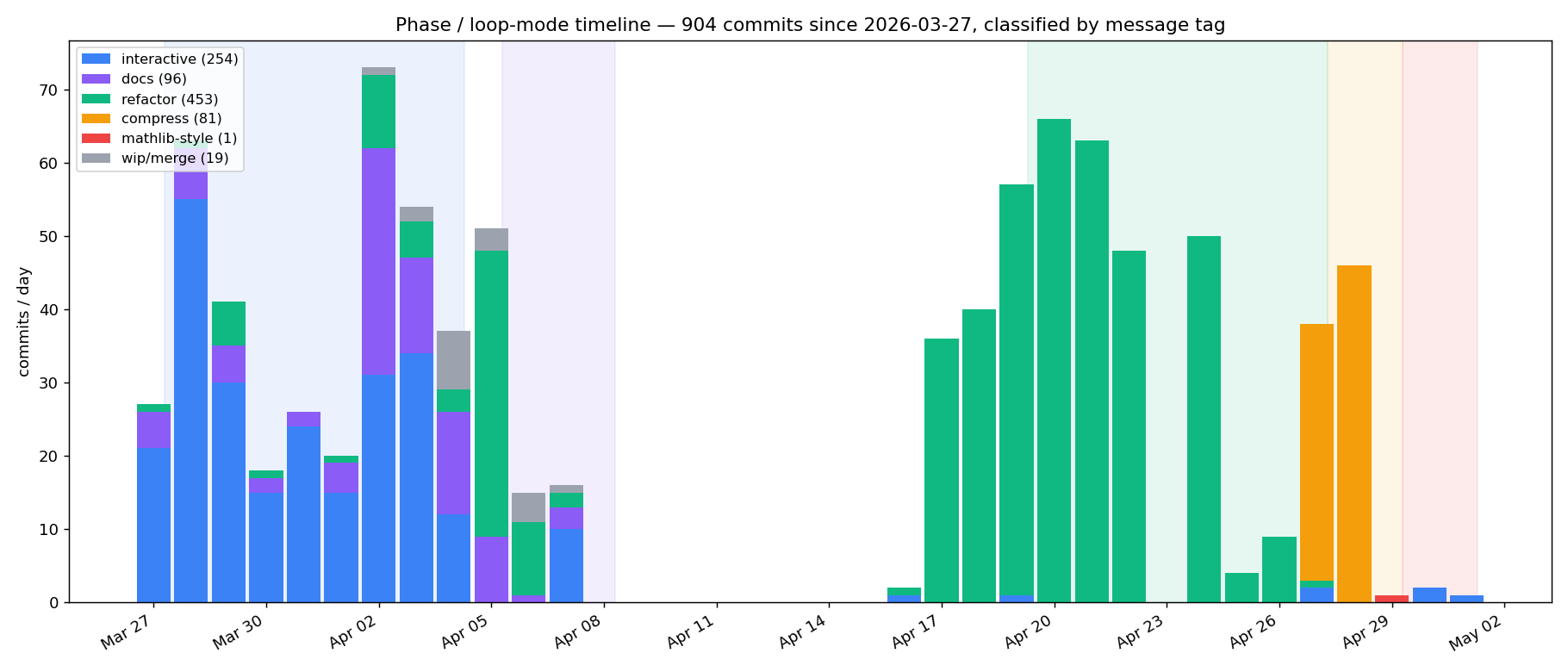}
  \caption{Daily commit counts classified by loop mode (commit-message
  tag).  Phase shading matches \cref{fig:timeline}.  Interactive commits
  dominate the proving phase (the formalization was not fully automated); the April~8--17 gap is the expert-review
  window; the refactor loop (green) drives the April~17--26 activity; the
  compression loop (orange) and a final mathlib-style polish (red) close
  out the project.}
  \label{fig:phase}
\end{figure*}

\begin{figure*}[t]
  \centering
  \includegraphics[width=.95\textwidth]{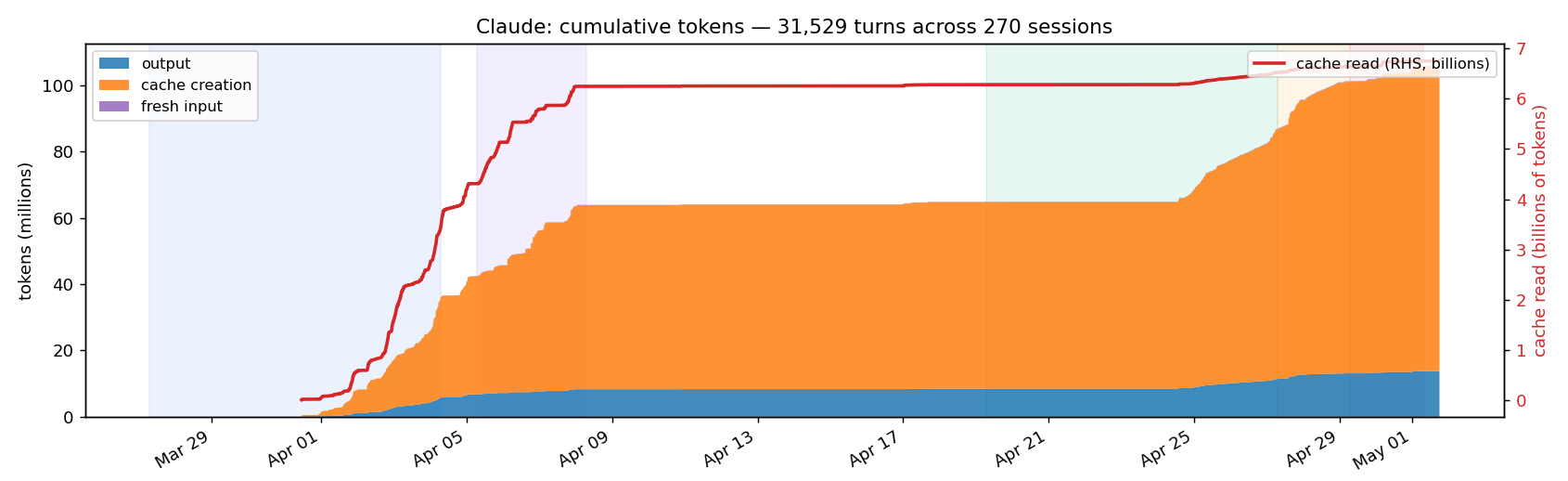}
  \caption{Cumulative Claude token usage across 31{,}529 turns in 270
  sessions.  Output (blue), fresh input (purple), and cache-creation
  (orange) are on the left axis (millions); cache reads (red) on the right
  axis (billions).  Cache reads exceed output by roughly two orders of
  magnitude: formalization is context-dominated because each step reloads
  \Lean{} files, diagnostics, and nearby library code.  At the posted Opus
  API rates (\$75/M output, \$18.75/M cache write, \$1.50/M cache read,
  \$15/M fresh input), the captured Claude usage corresponds to a total of
  roughly \$13K, dominated by cache reads (about \$10K). The authors used the \$200 subscription.}
  \label{fig:tokens}
\end{figure*}

\begin{figure*}[t]
  \centering
  \includegraphics[width=\textwidth]{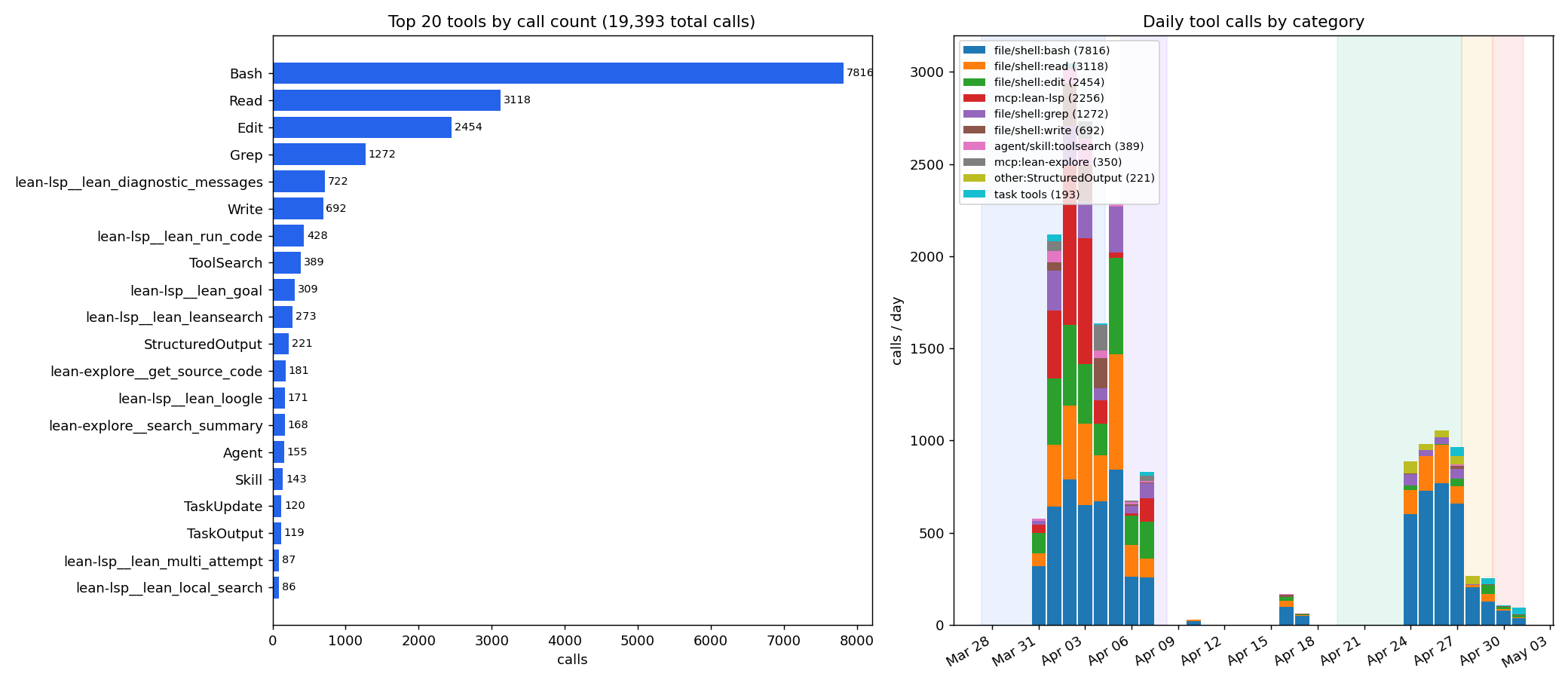}
  \caption{Tool calls across the project (19{,}393 total).  Left: top 20
  tools by call count, dominated by shell/file operations (\code{Bash},
  \code{Read}, \code{Edit}, \code{Grep}) and \Lean{} LSP queries.  Right:
  daily tool calls grouped by category, phase shading as in
  \cref{fig:timeline}.  Proof construction leans on \Lean{} LSP
  diagnostics, goal queries, and library search; the review response is
  dominated by repository edits, shell checks, and grep-style audits ---
  visibly different workflows on the same \Lean{} tree.}
  \label{fig:tools}
\end{figure*}

\paragraph{After the proof closed.}
As seen in \cref{fig:timeline} and \cref{tab:phase-timeline}, the project did not end when the sorry count reached zero. Instead, one of the authors performed a comprehensive review of the initial formalization (one week), which was converted into a checklist, and then the agent (Claude and Codex) iterated until the checklist was finished (two weeks). Three additional views of the same project (\cref{fig:phase,fig:tokens,fig:tools}) record commit
counts by loop mode, token usage, and tool-use mix; we refer to them
where relevant below. We refer to the initial formalization as state A, and the formalization that followed the first review as state B.

\section{Analysis}

\paragraph{What state A got right.}
The initial state had the proof in the right mathematical
shape. The overall structure of the proof was exactly as expected and followed the proof outline described above. While the goal of this article is to provide analysis of the quality of LLM-produced Lean code, we still need to note that it is remarkable that it is able to produce a proof this sophisticated with no human guidance other than the theorem statement and a standard textbook proof -- this was not possible one year ago.

Now that we have a formalized proof, the next question is: what can be done with the 5{,}000 lines of code that were produced?  Can the definitions and lemmas proved along the way to this theorem be repurposed and built off of?  These questions are the focus of our analysis.

Lean code, in the context of theorem proving, has three main components: definitions, theorem statements, and proofs. We report on how the agent did in both state A and state B for each of these.

\begin{table*}[t]
\caption{Qualitative before-and-after summary from the expert reviews.  The
table separates review items that have a local syntactic target from items that
require global API judgment.}
\label{tab:qual}
\centering
\small
\begin{tabular}{L{0.19\textwidth}L{0.34\textwidth}L{0.38\textwidth}}
\toprule
Review theme & State A criticism & State B outcome \\
\midrule
File structure &
File names were confusing and unorganized. &
Fixed. File names make sense and docstrings became readable. \\
Definitions &
Dozens of specific and often unnecessary definitions. &
Still the weakest category, did not noticeably improve after the review. \\
Theorem statements &
Most intermediate results were not general enough to be reusable. Agent proved exactly what it needed and nothing more. &
All specific changes requested in the review were done but the agent failed to identify further changes on its own. \\
API design &
Proofs worked by repeatedly unfolding definitions rather than building an interface. &
Partially fixed.  Downstream files are now cleaner but the API itself is noisy and bloated. \\
Proof style &
Proofs have long walls of \code{have} statements and frequent misuse of definitional equality. &
Better but uneven.  Some files became maintainable. The filtered-colimit files are still quite bad. \\
\bottomrule
\end{tabular}
\end{table*}

\paragraph{Definitions were the weakest point.}
Of the three main components to Lean code, definitions are by far the most important to get right and are where the agent struggled the most.

Bad definitions create future transport costs.  If a construction is placed at the
wrong level of generality, later users must compare it
to the version they actually need and then transport every relevant theorem
across that comparison. This transport can often be harder than simply redoing everything from scratch. When that is the case, it begs the question of whether there is any value in a formalization if it cannot be built upon effectively in the future. A proof can be ugly and still mostly function while a bad definition can create serious problems down the line.

The finitely generated subsheaf construction is a representative example.  The agent
defined finitely generated subsheaves of sheaves of abelian groups on a
topological space.  That was enough for the Grothendieck-vanishing proof, but the construction is not inherently about abelian
groups or even topological spaces. In the future, one may want finitely generated subsheaves for sheaves
of rings, or on sheaves over a site.  A human library designer would first design the
general construction and then provide convenient specializations.  The agent
created exactly what the current proof needed instead.

The same pattern appears in smaller definitions.  Some declarations merely
name a bijection that is needed once in a proof; others wrap a standard library
construction under a name that mentions the current application rather than the
mathematical object.  These choices are cheap in the moment because they give
the agent a handle to continue the proof.  They are expensive later because
they create public surface area whose intended use is unclear. In state A, there was a definition that roughly stated ``n is less than the Krull dimension of X''. This is mathematically identical to just writing \code{n < topologicalKrullDim X}. Not only do useless definitions like this make code more confusing, but they also hinder the ability of tactics like \code{simp}. These tactics use the syntax of the goal to decide on their next step so a useless definition like this will cause it to fail where it otherwise would have succeeded.

In total the agent made 62 of its own definitions and there was exactly one that was done the correct way, namely \code{TopCat.closedIncl}. This is a construction that exists in \mathlib{} for open sets but was missing for closed sets. The agent was able to build a useful API, likely copying the one from open sets in \mathlib{}. This is the kind of definition a library can keep,
the object is mathematically natural, the name describes the object correctly, and the adjacent API lets later proofs use it without needing to unfold its definition. The remaining 61 definitions were all done poorly and are not very useful for future formalizations. We give more specific examples of the problems with its definitions in a section below.

The agent did not improve its ability to make good definitions between state A and state B. Definitions remain the largest hurdle between the agent and writing usable, extensible code.

\paragraph{The API improved but not by much.}
After one decides on the right definition for an object, the next step is to build an API for interacting with it. A good test of the agent's ability to do this came with cohomology. The version of mathlib that the agent had access to possessed a definition of sheaf cohomology (\code{Sheaf.H}) but there had yet to be an API built for it. The first review complained that the agent did not build an API for cohomology at all and instead just repeatedly unfolded the definition. A specific change requested in the first review was for the agent to build an API for \code{Sheaf.H} and to not unfold its definition in downstream files.  In state B, the agent did create a dedicated cohomology API
file, and this did substantially improve the rest of the proof. While in state A, many files were littered with specific and ad hoc proofs about cohomology, in state B they were contained to one file and the proofs in later files became much cleaner. Unfortunately, the API itself was built very poorly.  Whenever a later proof needed a fact
about \code{Sheaf.H}, the agent simply added a new lemma to the file.  Most of these lemmas were highly specific to whatever proof the agent was writing at the time. In the end, the file totaled almost 800 lines of code with 24 lemmas in the documentation header. A human designing this would instead try and distill out the most important properties to design a small principled interface. This not only results in fewer lines of code, but more importantly, is much easier for future users to build off of.

The reviewer also requested the agent build an API for \code{topologicalKrullDim}. This was also defined in \mathlib{} but did not have an API for it yet. This definition is much simpler than that of cohomology and the agent did a much better job with it. It is still not at the level of \mathlib{} quality but it is much closer than the majority of machine-generated code. This is a good sign that the agent is capable of good design given sufficiently specific prompting.

\paragraph{Theorem statements improved more reliably.}
The response was much better at fixing local problems. The review had a number of specific requests for generalizations of theorems. In particular, state A had a number of theorems that had the hypothesis ``let S be a short exact sequence coming from an injective presentation'' while the statements were all true under the simpler hypothesis ``let S be a short exact sequence''. The agent had no problem making these changes.

In state A, there were a number of cases where the agent tried to prove ``$x = y$'', then realized it would be easier to just prove ``if $x = 0$ then $y = 0$'' since that turned out to be all it needed for the rest of the proof. This method of taking the path of least resistance makes sense when generating sorry-free code is the only goal but does not lead to useful, reusable code. This was pointed out in the review and it was fixed in state B. Even in cases where this change required a large refactor, the agent was able to fix them.

The review also prompted the agent to try and identify where these kinds of generalizations would make its code more useful and cleaner. Aside from a few artificial changes, the agent was not able to identify and correct any of these generality mistakes without them being requested. Much like with the definitions problems, the agent struggles to make decisions that have a global effect on the project and is much better suited for dealing with local problems.

Lean code of the type produced in this project, with an excess of specific lemmas and definitions, can still be useful to a search-heavy repository
of generated results.  In such a repository, an
agent can sift through excess statements and find a useful fact.  It is not useful to a human curated library like
\mathlib{}, where users need a smaller and more intentional theorem surface.

\paragraph{Proofs improved, but unevenly.}
The proof terms in state B are better than in state A. There are fewer walls of
intermediate \code{have} statements, less definitional-equality abuse, more
named intermediate lemmas replacing long unreadable proofs, and less direct unfolding of definitions. In general, the proofs that only used objects defined in \mathlib{} were much cleaner, as the agent was able to rely on the good API built in \mathlib{}. The proofs that used definitions made by the agent were extremely bad in state A and were noticeably better in state B after the APIs for Krull dimension and cohomology were built. Some of the more technical proofs remained bad, for example, the filtered-colimit files
remained hard to read because they contain a dozen hyper-specific definitions that are only useful in this one proof.

In some ways, the proof terms themselves are the least important part, as they are erased at runtime. There are performance and maintainability concerns however. Long tactic
blocks can create performance failures, and fragile definitional-equality
arguments are sensitive to breaks from changes upstream.  The project history
shows a recurring pattern: when a proof became too expensive to check, progress
improved only after the goal was decomposed into named sublemmas rather than
pushed through a larger heartbeat allowance.  That lesson is not specific to
this theorem.  It is the software-engineering version of a mathematical norm:
if a proof has a meaningful intermediate claim, name it and make later
maintenance depend on that claim.

\paragraph{More definitions as examples.}
As mentioned above, definitions are the most important thing to get right and are where the agent struggled the most. We will give some specific examples of the kinds of mistakes the agent made that it was unable to correct even after being prompted.

The worst offender is the definition \codesmall{sheafH_filtered_colimit_h1_sectionsFunctor}. This definition is equivalent to \code{sheafSections}, a definition available in \mathlib{}, and has nothing to do with sheafH, filtered colimits, or H1. The naming is likely because of the particular use case the agent had for this definition. Creating superfluous definitions like this is not just unnecessary, it can be actively harmful to future development. For example, if an instance is put onto this definition, say an instance of it being an additive functor, then even if \codesmall{sheafH_filtered_colimit_h1_sectionsFunctor} is shown to be equal to \code{sheafSections} (which it is), the instances will likely not be definitionally equal. This leads to some of the most confusing errors in Lean, where you can rewrite \codesmall{sheafH_filtered_colimit_h1_sectionsFunctor} into \code{sheafSections} but since the instances don't agree definitionally, everything will break.

Some other bad examples came from equivalences the agent defined. The definitions \codesmall{extClass_postcompAddEquiv_of_subsingleton_middle},
\codesmall{sheafH_extClassAddEquiv_of_subsingleton_middle}, and
\codesmall{sheafH_succ_iso_of_subsingleton_middle} are all roughly ``in this hyper-specific case, this function is bijective, so here it is as an \code{Equiv}''. The correct thing to do here is just prove the function is bijective and use that. Anytime an equiv is needed, \code{Equiv.ofBijective} can be inlined. The problem with making them definitions is that this essentially creates a barrier between the user and the actual function they should be working with, with no real upside.

\code{sheafH0EquivSections} and \code{sheafH0NatIsoSections} are essentially the same definition under different names, it should be clear why this is a bad thing.

In some cases, the agent made a choice that clearly went against previous design decisions. Sheaf cohomology is defined as a type, as opposed to a term of \code{AddCommGrpCat}. Both choices have their benefits, but choosing to go with the former was a deliberate design decision informed by considering all future applications. The agent chose to define \code{sheafH_succ_map} as a morphism between the \code{AddCommGrpCat} versions of cohomology. This is not a bad definition in principle but it clearly goes against a previous design decision and working with both adds work that is not needed.

Some definitions, like \code{familyMap} and \code{familyImage} are just silly. The definition of \code{familyMap f} is just \code{Sigma.desc f} so this definition saves exactly one character of typing and does nothing else. Similarly, \codesmall{TopologicalSpace.IrreducibleCloseds.height} is just defined to be \code{Order.height}. This one has some use, as it allows the dot notation when working with irreducible closeds. However, because it chose to make it a \code{def} instead of an \code{abbrev}, all of the theorems in \mathlib{} proven about \code{Order.height} will not be usable with it unless it is explicitly unfolded, which is exactly what the agent does when it uses it.

\paragraph{Why feedback adaptation is uneven.}
The successful review items had crisp completion predicates: no remaining uses
of an old name, a file renamed, a theorem statement generalized, a wrapper
deleted, or a stronger isomorphism available at a call site.  The failed items
required counterfactual judgment.  A good definition is good because future
developments can use it without transport pain.  A good API is good because it
contains the right small set of lemmas, including lemmas whose need has not yet
appeared.  The agent optimized for the next compiling proof, but was unable to make good long-term decisions.

\paragraph{Prompts had to forbid escape hatches.}
A recurring failure mode during long autonomous runs was loop drift toward
declaring a task ``blocked'' or pinning a failure on infrastructure
(``genuine mathlib gap'').  The drift was fast and consistent enough that
the project's slash-command prompts explicitly forbade both
phrases --- ``You are NOT allowed to say that you are blocked. You MUST
close the sorrys yourself!'' and ``A no-op cycle is never acceptable.''
When an agent has a verbal escape hatch, it will reach for it before it reaches for a decomposition; the gate partially removes the escape hatch.

\paragraph{Human feedback as a specification.}
The review supplied the quality specification the automated loop
otherwise lacked.  ``Build a useful cohomology API'' reads like a
programming task but is really a design task, and a loop can satisfy
every local complaint while still producing an interface that is too
broad, too specific, or too hard to navigate.  The division of labor
follows: expert reviewers should spend their scarce attention on
definitions, theorem surfaces, and global organization, and hand agents
the narrow refactors that carry an explicit completion predicate.

\paragraph{Gates are only as good as their target.}
The two loops shared the worker/evaluator structure but had different
reward signals.  The compression loop reduced normalized LOC under a
compilation gate: a clean mechanical signal, which made the loop a
reliable optimizer but only loosely aligned with review intent.  The
refactor loop had to operationalize natural-language review items, and
its evaluator could report progress as strong, partial, or absent
without ever being sure a broad review item was truly satisfied.  The
lesson is that a mechanical gate is powerful when it matches the quality
target and dangerous when it is only a proxy.  Compilation gates are
correctness gates; LOC and dead-code gates are maintenance proxies;
review-item checklists are alignment proxies.  None of them measures
whether the public definitions are the ones a future formalizer would
choose --- exactly the gap between state A and the first review.

\paragraph{External proving as a bounded tool.}
Aristotle submissions are local assistance, not an autonomous strategy
engine.  Successful calls closed bounded lemmas; disproven calls were
informative because they falsified incorrect lemmas.  But the hard
project decisions --- which intermediate statements should exist, which
definitions should be public --- were never theorem-prover queries.

\section{Conclusion}

Without a sophisticated harness, LLMs are very capable of closing proofs and local changes, but cannot
yet make the global design decisions that a reusable formalization
requires.  Choosing good definitions and building a good API remain the
biggest barriers.  We suggest a practical evaluation standard for
AI-for-math systems: after the proof checks, ask an expert to review
the public definitions, theorem statements, namespaces, and API
surface, then measure how well the system responds.

\section*{Code and Data Availability}

The \Lean{} 4 formalization is available at
\url{https://github.com/Vilin97/Clawristotle/tree/main/grothendieck-vanishing}.  The two expert reviews and
the structured process logs analyzed in this paper---per-turn token usage,
the refactor and compression loop histories, per-commit LOC and sorry
counts, the agent tool-use timeline, Aristotle proving jobs, the human
prompts, and the loop prompts---are released as a public dataset at
\url{https://huggingface.co/datasets/uw-math-ai/grothendieck-vanishing-logs}.

\section*{Acknowledgements}

This work used computational resources from the Hyak high-performance
computing cluster at the University of Washington. The AI subscription was funded by the UW Math AI Lab. We thank the
\mathlib{} community, whose library and conventions defined the standard of
quality against which this formalization was reviewed.

\clearpage
\bibliography{references}
\bibliographystyle{icml2026}

\clearpage
\appendix

\section{Additional Figures}

This appendix collects diagnostic material that helps audit the case study
but is too detailed for the main argument: a per-cycle view of loop
effectiveness, and the final import graph of the \Lean{} development.

\paragraph{Loop effectiveness.}
The loop-effectiveness plot gives a per-cycle view of the refactor and
compression loops.
Refactor cycles are scored by progress and task-completion status, compression
cycles by LOC delta, which is why the two loops should not be compared as if
they shared an objective: one was interpreting a natural-language code review,
the other optimizing a mechanical signal.  The plot also shows why the paper
does not treat cycle count as evidence of quality by itself --- some stretches
are productive, others are repeated low-value attempts.

\begin{figure}[h]
  \centering
  \includegraphics[width=\linewidth]{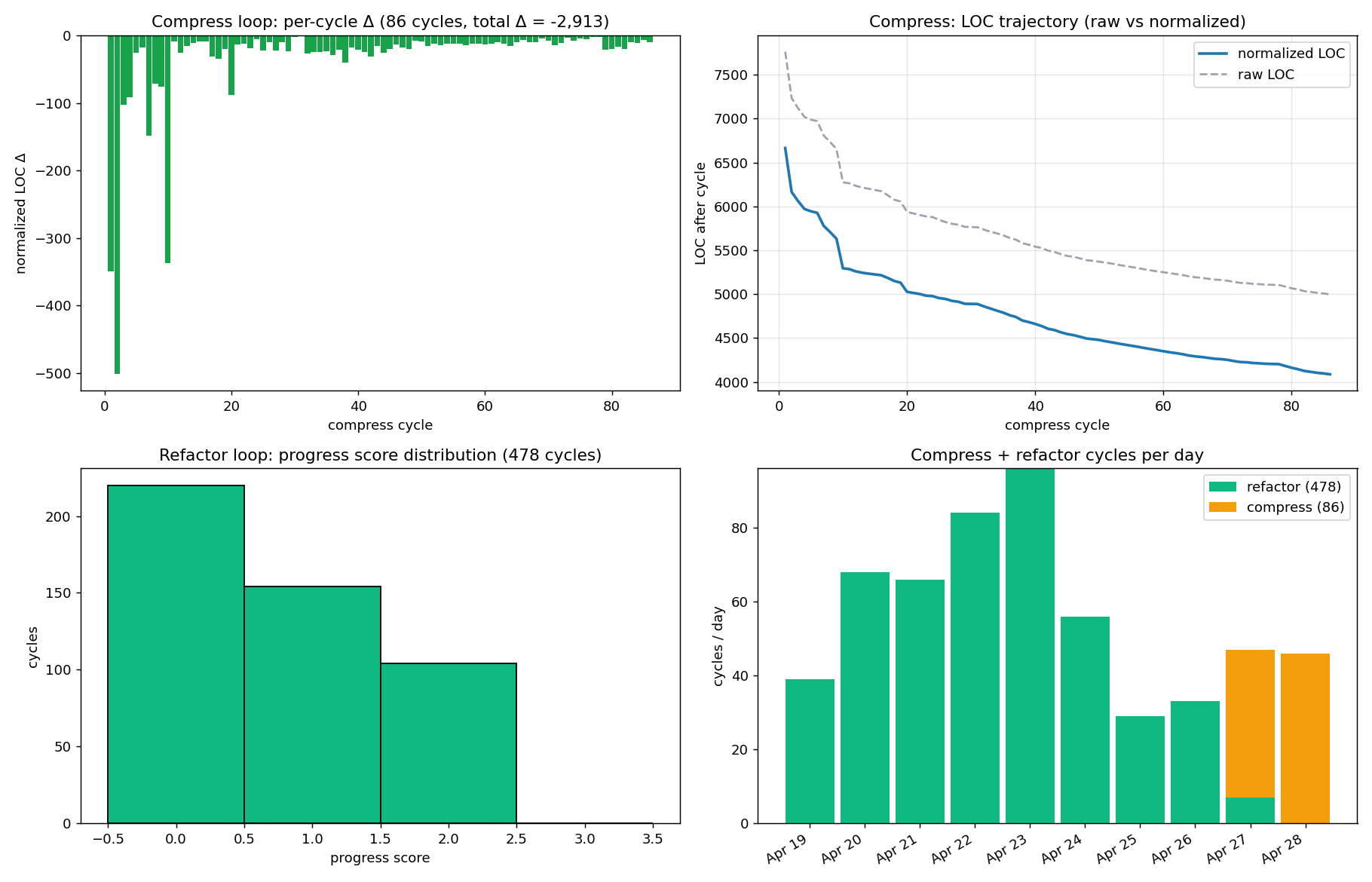}
  \caption{Per-cycle refactor and compression outcomes summarizing loop
  behavior across the project.}
\end{figure}

\paragraph{Import graph.}
The import graph records the final organization of the \Lean{} development.
Many review complaints were about where concepts lived rather than about
isolated proof terms, so the organization is itself a result.  The intended
reading is bottom-up: dimension, zero-outside, flasque, closed-immersion,
filtered-colimit, and generated-subsheaf infrastructure feed the irreducible
step and then the main theorem.  A future user navigates this artifact through
imports and namespaces before reading any proof, which is why file naming
carried weight in the review.

\begin{figure}[h]
  \centering
  \includegraphics[width=\linewidth]{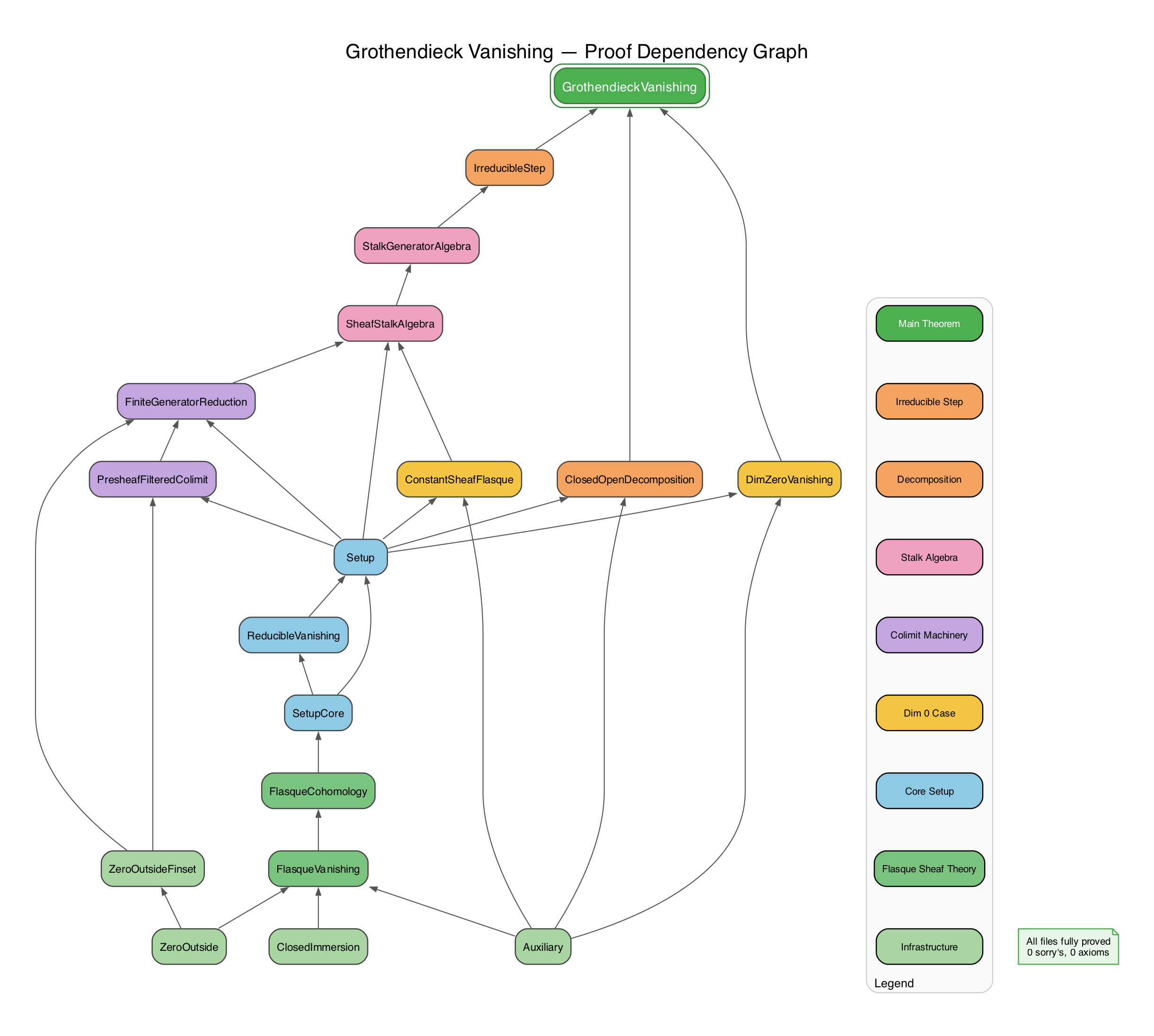}
  \caption{Import graph for the Grothendieck-vanishing Lean files.}
\end{figure}

\end{document}